\documentclass[conference]{IEEEtran}
\IEEEoverridecommandlockouts

\usepackage{cite}
\usepackage{amsmath,amssymb,amsfonts}
\usepackage{algorithmic}
\usepackage{graphicx}
\usepackage{textcomp}
\usepackage{xcolor}
\usepackage{float}
\usepackage{braket}
\usepackage{booktabs}
\usepackage{subcaption}
\usepackage{hyperref}
\def\BibTeX{{\rm B\kern-.05em{\sc i\kern-.025em b}\kern-.08em
    T\kern-.1667em\lower.7ex\hbox{E}\kern-.125emX}}
\begin{document}

\title{Quantum Models with Multi-Stage Training for Compositional Concept Generalization\\
\begin{center}
\footnotesize
\parbox{0.95\textwidth}{
\centering
 Accepted for publication at IEEE QCE 2026 (IEEE Quantum Week 2026). Open Access, CC BY.}
\end{center}
\vspace{-0.8em}
}

\author{
\IEEEauthorblockN{Mina Abbaszadeh}
\IEEEauthorblockA{
Department of Computer Science \\
University College London \\
London, United Kingdom \\
m.abbaszadeh@ucl.ac.uk
}
\and
\IEEEauthorblockN{Matilda Karabina Moore}
\IEEEauthorblockA{
Department of Computer Science \\
University College London \\
London, United Kingdom \\
matilda.moore.21@ucl.ac.uk
}
\and
\IEEEauthorblockN{Mehrnoosh Sadrzadeh}
\IEEEauthorblockA{
Department of Computer Science \\
University College London \\
London, United Kingdom \\
m.sadrzadeh@ucl.ac.uk
}
\and
\IEEEauthorblockN{Martha Lewis}
\IEEEauthorblockA{
Institute for Logic, Language and Computation \\
University of Amsterdam \\
Amsterdam, Netherlands \\
m.a.f.lewis@uva.nl
}
}

\maketitle

\begin{abstract}

Compositional Concept Generalization (CoCoGen), the ability to systematically recombine learned primitives in novel contexts, is a key challenge for multimodal learning. In this work, we provide a solution  using a compositional model of meaning that separates nouns from relations and uses tensors and variational quantum circuits to train them on data. This model enables us to employ a multi-stage training paradigm, one that first learns object representations from single-object image–caption pairs, then subsequently transfers these to the relational stage where object parameters are frozen and optimisation is only applied to relational components. This design explicitly enforces compositional factorisation at the circuit, ensuring that relations are learned as transformations over stable primitives.
The training paradigm is tested on the CLEVR dataset developed specificially for CoCoGen. For text, we work with vector representations of nouns and higher order tensor representations of relations using a set of different ansatz. For images, we work with quantum encodings of image embeddings dervied from Open AI's Vision-Language tool CLIP and contrast amplitude encoding, which preserves the original embedding geometry, with angle encoding, which introduces nonlinear feature transformations.
Our results show that multi-staged training combined with structured encodings significantly improves out-of-distribution relational generalisation, while using orders of magnitude fewer trainable parameters than classical baselines. We find that performance gains arise from the interaction between representation and encoding, with nonlinear quantum encodings enhancing the separability of compositional structure.
These findings demonstrate that structured quantum representations and staged learning provide an effective framework for compositional generalisation in multimodal quantum machine learning.

\end{abstract}


\section{Introduction}
Humans can understand novel situations by recombining previously learned concepts, a capability known as \emph{Compositional Concept Generalization} abbreviated to CoCoGen. Consider the situation where  a human  has learned the concept of cars but has never seen a yellow car. Now if they see a yellow car when  crossing a road, they will still be cautious, knowing it is dangerous. Achieving this ability in current AI models remains an open challenge, particularly for learning relational concepts that must generalize beyond seen combinations \cite{thrush2022winoground,hsieh2023sugarcrepe,Lewis2023,pearson2025}. In principle, AI models that are based on compositional models of meaning and which learn using the principles of compositionality should be able to generalize concepts appropriately. An example of such a model is the  Distributional Compositional Categorical  model (DisCoCat) \cite{Coecke2010,grefenstette-sadrzadeh},  which offers a framework for compositional language understanding. DisCoCat represents word meanings by higher-order tensors that are composed via tensor contraction,  according to linguistic structure, i.e. the rules of syntax. Although theoretically well suited for compositional generalization, DisCoCat has so far shown limited empirical success in experiments, in part due to the computational cost of learning and manipulating higher-order tensors on classical hardware \cite{lewis2022clipbinding}. 

 DisCoCat shares its underlying mathematical underpinnings with Categorical Quantum Mechanics (CQM)  and a circuit optimisation framework known as the ZX-Calculus~\cite{abramskybob2004,Duncan2020ZX,coecke2020quantum,CoeckeKissinger2017}. This fundamental structural link allows one to translate DisCoCat representations to Variational Quantum Circuits (VQCs), then train them by running simulations on classical computers, which has led to a novel field of research known as Quantum NLP~\cite{lorenz2021qnli,kartsaklis2021lambeq,Liu2023DisCoCirc}. This capability  significantly improves DisCoCat's trainability but has only been tested on natural language  tasks, such as classification see~\cite{Lorenz2021,Meichanetzidis_2023,WazniLoPheatSadr2024},  interpretability~\cite{Tull2024Interpretability} and logical reasoning~\cite{duneau2024}. In this paper, we use this capability to apply them to compositional generalisation, which is a multimodal learning task. The use of quantum circuits in this work is motivated by their natural compatibility with the compositional structures used by DisCoCat. 
 
 The DisCoCat framework is defined in terms of tensor-based compositional representations, which can be directly mapped to quantum circuits. Nevertheless, this is not merely an implementation choice: quantum circuits natively realise tensors, enabling composition through rotation and controlled rotation gates and circuit connectivity. In addition, quantum circuit representations offer a compact parameterisation of high-dimensional tensor spaces, allowing structured interactions between components (e.g., objects and relations) to be modelled efficiently. In contrast, classical tensor-based implementations learn the tensors as multi dimensional arrays and  require substantially larger parameters to achieve comparable results~\cite{maillard-clark,wijnholds-clark-sadr}.

Despite the above advances, DisCoCat needs to be  grounded in multimodal data to model compositional concept generalization. Multimodal extensions of DisCoCat are  recent and limited in focus, e.g. see 
\cite{nazir2024adjective}, 
for an extension to adjective-noun composition in audio-text data.  At the same time, vision--language models such as CLIP \cite{radford_learning_2021} have achieved strong performance across multimodal benchmarks but exhibited limited compositional generalization, particularly in binding objects to relations \cite{Lewis2023,pearson2025}. Previous work explores extensions of DisCoCat to verbs~\cite{verbclip,discoclip} but firstly only train the tensors classically, also does not consider CoCoGen tasks and focuses on general purpose text-image alignment.  We are therefore filling this gap by exploring the ability of quantum models in achieving CoCoGen via the compositional models of semantics leading to structured VQCs. The contributions of this work are two-fold:
\begin{enumerate}
    \item Evaluating one such compositional model, i.e. DisCoCat and its translation VQCs in a controlled compositional generalization benchmark involving geometric objects and spatial relations.
    \item Proposing a \emph{multi-stage training paradigm} that disentangles object  from relational learning. This staged training strategy is a form of curriculum learning, where the model first learns simple object-level representations before progressing to more complex relational reasoning. Unlike standard curriculum learning, where parameters are continuously updated across stages, we explicitly preserve learned object representations by freezing parameters, thereby enforcing compositional factorisation between objects and relations.
\end{enumerate}

The results show that DisCoCat-based VQCs achieve stronger relational generalization under out-of-distribution (OOD) splits than classical baselines, notably OpenAI's CLIP model. Furthermore, multi-stage training consistently outperforms single-stage variants, highlighting the importance of separating object grounding from relational learning. In particular, these improvements are achieved using very substantially fewer trainable parameters (426 vs 151 million). Our focus in this work is on the representational and compositional properties of quantum models, rather than hardware-level performance. Accordingly, we evaluate our approach in simulation, which allows us to isolate the effect of model structure independently of noise and device-specific constraints.

\begin{figure*}[t]
   \centering
   \includegraphics[width=\linewidth]{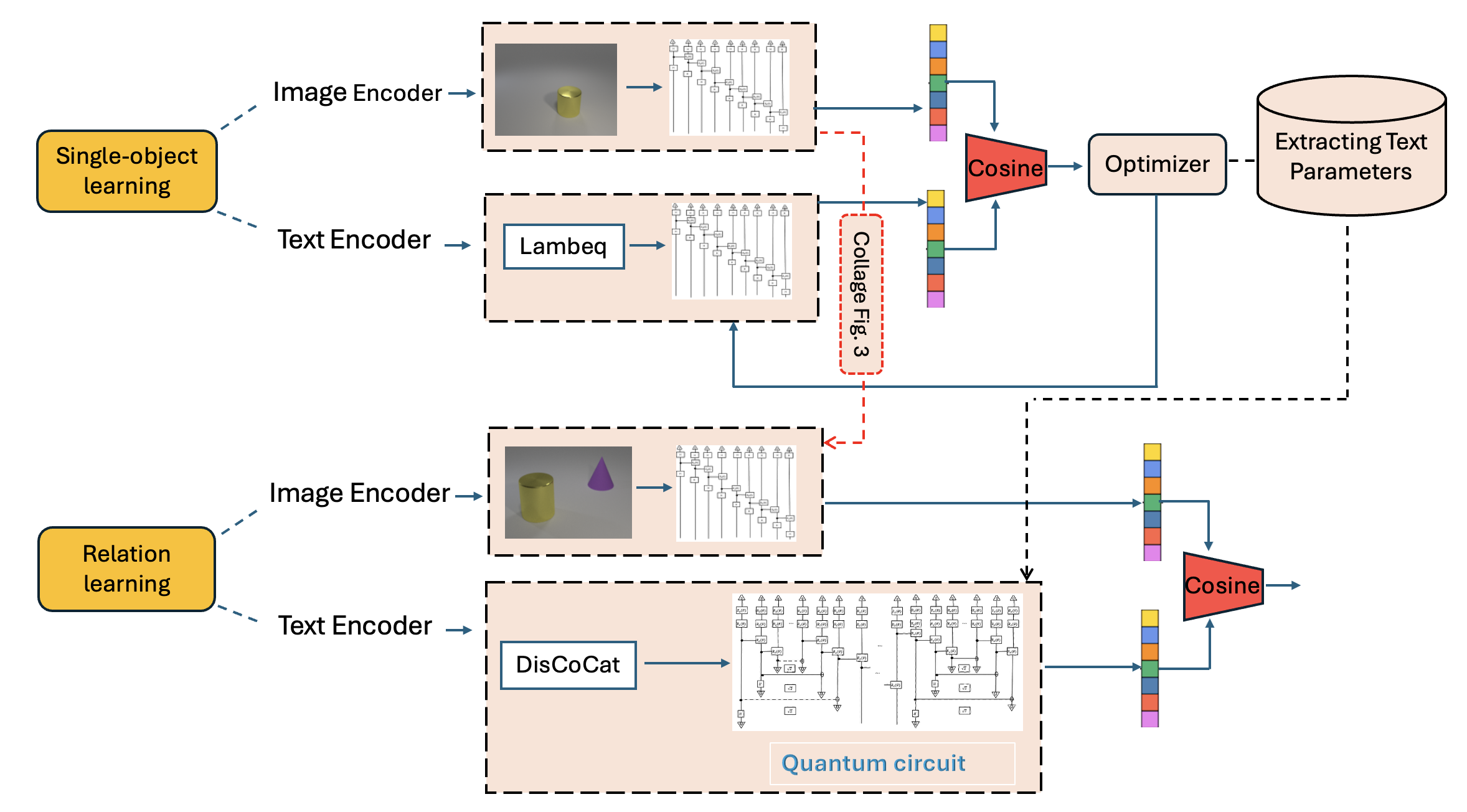}
   \caption{A multi-stage procedure for compositional image–caption alignment. After optimizing the noun parameters during single-object training, the learned noun parameters are transferred to relational training. In the collage image encoding, relational image circuits are constructed by combining the quantum circuits of single-object images.}
   \label{pipeline}
\end{figure*}

\section{Background}
In this section, we briefly review the theoretical foundations underlying DisCoCat, its translation to variational quantum circuits, and relevant multimodal baselines.
\subsection{DisCoCat Meaning Representations in Hilbert Space}
\label{sec:Disco}

The basic idea behind DisCoCat is to model  the grammar and the meaning  of a sentence  in two different compact closed categories (CCC), and then set up a functor from the grammar category to the meaning category. 
In this way, the compact closed structure is preserved and the rules of language defined in the grammar category are interpreted as morphisms in the meaning category, which enable composition of word representations. 
A CCC is a  monoidal category  $(C,\otimes, I)$ such that for each object $A\in C$ there are objects $A^l,A^r\in C$ (the \emph{left} and \emph{right duals} of $A$) and the following morphisms (satisfying certain conditions, see here \cite{Coecke2010,preller2011bell}

\begin{equation}
\begin{aligned}
\eta_A^l &: I \rightarrow A \otimes A^l, \qquad
\eta_A^r &: I \rightarrow A^r \otimes A, \\
\epsilon_A^l &: A^l \otimes A \rightarrow I, \qquad
\epsilon_A^r &: A \otimes A^r \rightarrow I .
\end{aligned}
\end{equation}

\textbf{Grammar} is formalized using a partially ordered \emph{pregroup}, which is a tuple $(A,\cdot , 1, \--^l,\--^r,\leq )$ where $(A,\cdot, 1, \leq)$ is a partially ordered monoid and $\--^r, \--^l$ are functions $A\rightarrow A$ such that $\forall x\in A$,  we have 
\begin{equation}
    x\cdot x^r \leq 1 \leq x^r \cdot x \qquad \qquad  \qquad \qquad  x^l\cdot x\leq 1 \leq x \cdot x^l
    \label{ref:prgeq}
\end{equation}
providing the duals and morphisms of the CCC.
The binary $\cdot$ operation is usually omitted, writing $xy$ for $x\cdot y$.  The structure of a typical caption of our dataset is a sentence in the form: `$x$ is to the right/left of $y$'. 
In order to decrease the circuit complexity, we simplify the sentences to the form `noun $\{$isLeftOf/isRightOf$\}$ noun', which has the type $n \ (n^r s n^l) \ n \leq 1s1 = s.$
\textbf{Meaning} is formalized in finite dimensional Hilbert spaces and their tensor products. This forms a CCC, where  $I$ is the one-dimensional space $\mathbb{R}$, $A^r = A^l = A$,  the $\epsilon$ maps takes the inner product, and the $\eta$ map creates a diagonal matrix:

\begin{equation}
\begin{aligned}
\epsilon_V &: V \otimes V \rightarrow \mathbb{R}, \\
&\sum_{i,j} c_{i,j} (v_i \otimes v_j)
\mapsto \sum_{i,j} c_{i,j} \langle v_i \mid v_j \rangle, \\[4pt]
\eta_V &: \mathbb{R} \rightarrow V \otimes V, \\
&1 \mapsto \sum_i (v_i \otimes v_i).
\end{aligned}
\label{eq:fhilb_cups}
\end{equation}

\subsection{From DisCoCat to Variational Quantum Circuits (VQCs)}
The translation from CCC diagrams to quantum circuits depends on the choice of ansatz. 
For quantum--classical hybrid learning, a variety of ansatzes are available, including the Instantaneous Quantum Polynomial (IQP) ansatz~\cite{Havlicek2019QuantumFeatureSpaces}, the Sim family ansatz~\cite{Sim2019Expressibility}, and tensor-network-based constructions such as Matrix Product States (MPS)~\cite{stoudenmire2016supervised}. 
In order to control the number of qubits and trainable parameters while maintaining expressive power, we use the Sim4 ansatz to translate DisCoCat diagrams of captions into quantum circuits, and the IQP ansatz to encode image embedding vectors. Further implementation details are provided in Section~\ref{Quantum-Circuit-Implementation-Details}.

\subsection{CLIP and the Contrastive Learning Paradigm}
\label{sec:clip}
We use the CLIP (Contrastive Language-Image Pretraining) model \cite{radford_learning_2021} as a classical baseline as it is widely used throughout vision-language modelling. The architecture of CLIP is briefly described as follows. CLIP comprises two Transformer-based embedding models: one for image inputs and one for text inputs. These are called the \textit{image encoder} and the \textit{text encoder}. When given an input of an image paired with some text, the image encoder produces an \textit{embedding} (i.e., a vector) representing the image, and the text encoder produces an embedding representing the text. The parameters of the model are trained to align the embeddings of the images and the text using both positive and negative examples. That is, given a \textit{positive} image-caption pair, for example an image of a dog and a text caption `a photo of a dog', the model is trained to increase the cosine similarity (inner product of normalised vectors) of the two embeddings. Given a \textit{negative} image-caption pair, for example an image of a dog and a caption `a photo of an aeroplane', the model is trained to decrease the cosine similarity between the two embeddings, moving them apart in the space. The use of both positive and negative pairs for training is generally termed \emph{contrastive learning}.

In our experiments, we use CLIP image embeddings as input features for both classical and quantum models. We freeze these embeddings, meaning that we do not alter these inputs during training.
For comparison with the quantum text encoder, we implement a CLIP–Text Projection (CTP) baseline, in which a CLIP text encoder is fine-tuned jointly with a lightweight projection head under the same supervised contrastive objective used in our quantum models. 
This ensures that performance differences are attributable to representational structure rather than differences in training objectives. This setup provides a direct classical baseline, where similarity is computed between CLIP embeddings using cosine similarity, enabling comparison with the proposed quantum representations. 
Unlike the DisCoCat-based quantum models, CLIP does not explicitly encode compositional structure in its architecture. As such, it serves as a strong non-structural baseline for evaluating relational generalization. 

\section{Methods}\label{Methods}

\subsection{Task and Datasets}\label{Datasets and Task Formulation}
\begin{figure}[t]
   \centering
   \includegraphics[width=\linewidth]{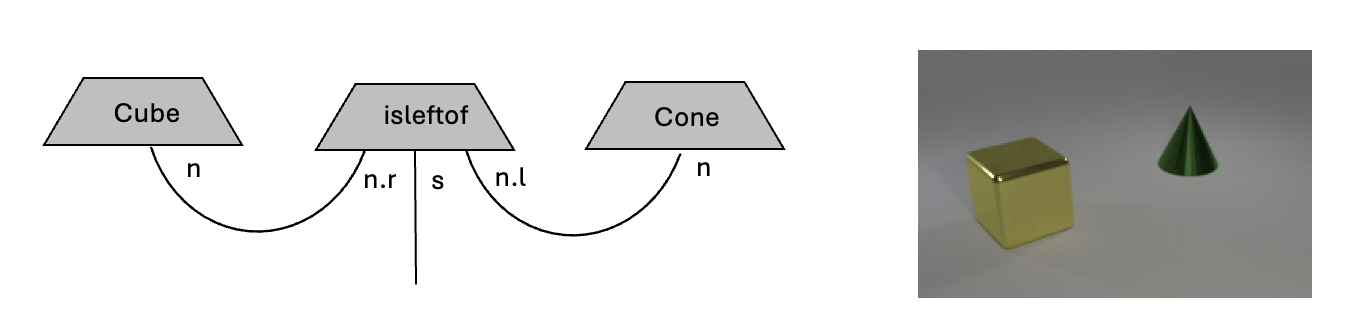}
   \caption{Example from the relational dataset with the correct caption ``cube left cone'' and the relation-swapped distractor ``cube right cone''. The corresponding DisCoCat diagram is shown on the left.}
   \label{cube-left-cone}
\end{figure}

The task is an image--caption alignment problem for compositional generalization. A model is trained to associate images of shapes in particular configurations with their corresponding captions, as illustrated in Figure~\ref{cube-left-cone}. The model is then evaluated on images whose corresponding captions represent compositions not observed during training. For example, a model trained on instances such as \textit{cone right cube} is expected to generalise to unseen combinations such as \textit{cone left cube}.
We use the single-object and relational splits of the CoBi2 dataset~\cite{pearson2025}. 
The single-object dataset contains images consisting of a single object, where the object can be one of four shapes (\emph{cube}, \emph{sphere}, \emph{cylinder}, \emph{cone}), with varying sizes, colors, and positions. The relational dataset consists of images containing two distinct shapes, with each image paired with one correct caption of the form \textit{shape$_1$ left/right shape$_2$} and one distractor obtained by swapping the relation \textit{left} $\leftrightarrow$ \textit{right}. 
The controlled scale of the dataset allows us to isolate compositional generalisation independently of perceptual complexity. We emphasize that this benchmark is not introduced in this work, but is adopted from prior studies on compositional generalization in vision--language models~\cite{pearson2025}. Previous work has shown that even such minimal compositional settings remain challenging for modern vision--language models, particularly in out-of-distribution generalisation~\cite{hsieh2023sugarcrepe, thrush2022winoground, hupkes2020compositionality}. 
The controlled nature of the dataset is therefore intentional: it provides a well-established benchmark for evaluating compositional generalisation independently of perceptual complexity, allowing improvements to be attributed to the model architecture rather than dataset scale. This evaluation setting follows established practice in compositional generalisation research, where controlled benchmarks are used to rigorously assess the ability of models to recombine known components in out-of-distribution settings.
The dataset is divided into training, out-of-distribution (OOD) validation, and OOD test splits with no overlapping compositional configurations across splits. 
Given an image and two candidate captions, the task is to select the correct caption by assigning it a higher similarity score than the distractor. 
We emphasise that this is not a standard classification task but an image--caption alignment problem with distractors. Unlike classification, which does not require compositional reasoning, the alignment setting requires the model to distinguish between correct and incorrect compositional descriptions, including out-of-distribution combinations. This makes the task substantially more challenging.
In total, 24 possible relational captions are generated, each associated with multiple visually distinct images.

Table~\ref{tab:dataset_splits} summarizes the train, validation and test
splits used in our experiments following \cite{pearson2025}.
\begin{table}[h]
\caption{Dataset splits for the compositional generalisation benchmark~\cite{pearson2025}.}
\label{tab:dataset_splits}
\centering
\begin{tabular}{lccc}
\hline
Dataset & Train & OOD Val & OOD Test \\
\hline
Single-Object & 1360 & 400 & 1100 \\
Two-Object    & 7440 & 600 & 3700 \\
Relational    & 440  & 250 & 400 \\
\hline
\end{tabular}
\end{table}

The splits are constructed by withholding specific relational triples of the form (shape$_1$, relation, shape$_2$). While all individual shapes and relations are observed during training, only a subset of their compositions is included in the training set. The validation and test splits contain combinations that are not seen during training but are composed of previously observed primitives, requiring the model to recombine known objects and relations into novel configurations. This corresponds to a weak form of systematic generalisation, in which generalisation arises from recombination of known components rather than exposure to new primitives.

Figure~\ref{cube-left-cone} illustrates an example from the relational dataset, showing an image together with its correct caption.

\subsection{Two-Stage Training Procedure}\label{Two-Stage Training Procedure}
Compositional generalization requires stable conceptual primitives. In order to obtain these primitives, we use a two-stage training regime. We first train models to caption images from the single object dataset. We then use frozen parameters from these models as inputs for models trained to caption images from the relational datasets. 
In \textbf{Stage 1}, models are trained on single-object image–caption pairs (“cone”, “cube”, “sphere”, “cylinder”) using contrastive learning. This stage aligns embeddings of images of individual objects with embeddings of their corresponding text captions. The purpose of this stage is to learn stable object representations shared across modalities and to ensure that geometric shapes are learned independently of their relational context.
In \textbf{Stage~2}, we perform relational learning with frozen object parameters. Relational captions (e.g., \emph{cube left cone}) are introduced. Object parameters learned in Stage~1 are frozen, and only relational parameters (\emph{left}, \emph{right}) are trained. Training is formulated as a binary image–caption alignment task: for each image, a correct caption is paired with one relation-swapped distractor caption, and the model is trained to assign a higher score to the correct pair.
This training procedure can be interpreted as enforcing a factorised compositional structure, which we formalise in the following subsection.

\subsection{Compositional Factorisation via Multi-Stage Training}

A central requirement for compositional generalisation is the ability to represent structured inputs as compositions of reusable primitives. In the relational setting, this corresponds to modelling representations under a factorised compositional structure of the form
\begin{equation}
f(x, r, y) = \mathcal{C}\big(g(x),\, h(r),\, g(y)\big),
\end{equation}
where $x$ and $y$ denote objects, $r$ denotes a relation, $g(\cdot)$ is a representation function for objects, $h(\cdot)$ is a representation function for relations, and $\mathcal{C}(\cdot)$ denotes a composition operation. In DisCoCat-based models, this composition is implemented via tensor contraction according to grammatical structure, corresponding in the quantum setting to circuit composition and contraction of wires.
Our training procedure enforces this structure in a sequential manner. In Stage~1, we learn object representations
$g(x; \theta_{\mathrm{obj}})$
using single-object image--caption pairs, without introducing relational structure. This stage ensures that object representations are learned independently of relational context. In Stage~2, we construct relational representations by composing the learned object representations with a relation component:
\begin{equation}
f(x, r, y) = \mathcal{C}\big(g(x; \theta_{\mathrm{obj}}),\, h(r; \theta_{\mathrm{rel}}),\, g(y; \theta_{\mathrm{obj}})\big),
\end{equation}
where the object parameters $\theta_{\mathrm{obj}}$ are frozen and only the relation parameters $\theta_{\mathrm{rel}}$ are optimised.
This separation constrains the model to learn relations as transformations over fixed object representations, reducing the tendency of relational parameters to re-encode object-specific information. As a result, the model is encouraged to generalise to unseen compositions by recombining independently learned primitives.
In the quantum setting, this factorisation aligns naturally with the tensor product structure of Hilbert spaces, where compositional structure is represented through circuit composition. The multi-stage training procedure can therefore be interpreted as enforcing compositional structure at the level of quantum circuit parameters.
\subsection{Image Encoding}
\subsubsection{One-Hot Encoding (OHE) and Multi-Hot Encoding (MHE)}\label{MHE}
Firstly, as a proof of concept, we use one-hot (OHE) or multi-hot encoding (MHE) following \cite{ieee2025}, with five qubits per noun. 
Each image is encoded as a 5-qubit quantum circuit in the IQP ansatz, where the components of the vectors shown in Table \ref{tab:imageOHE-shape} are used as parameters of the entangling gates. For multi-hot encoding (MHE), these vectors are concatenated to encode relational images as shown in Table \ref{tab:imageMHE-preposition}.

\begin{table}[t]
\caption{MHE vectors for sentences shape.}
\label{tab:imageMHE-preposition}
\centering
   \hspace{-0.5cm} \begin{minipage}{7cm}
        \centering
        \begin{tabular}{lc}
            \hline
            \multicolumn{2}{c}{\textbf{Shapes}} \\
            \hline
            Cylinder & $[1, 0, 0, 0]$ \\
            Sphere   & $[0, 1, 0, 0]$ \\
            Cube     & $[0, 0, 1, 0]$ \\
            Cone     & $[0, 0, 0, 1]$ \\
            \hline
        \end{tabular}
        \caption{OHE vectors for noun shape.
        }
        \label{tab:imageOHE-shape}
    \end{minipage} \qquad  
    \begin{minipage}{7cm}
        \centering
        \begin{tabular}{lc}
            \hline
            \multicolumn{2}{c}{\textbf{Sentences}} \\
            \hline
             X Left-of Y   & $[\underbrace{a, b, c, d}_{\mbox{OHE of X}}, \underbrace{a', b', c', d'}_{\mbox{OHE of Y}}]$ \\
            X Right-of Y &$[\underbrace{a', b', c', d'}_{\mbox{OHE of Y}}, \underbrace{a, b, c, d}_{\mbox{OHE of X}}]$ \\
            \hline
        \end{tabular}
    \end{minipage}
\end{table}

\subsubsection{Quantum Encodings of CLIP Embeddings}

Secondly, we encode frozen CLIP embeddings of the images using quantum representations. CLIP embeddings have 512 dimensions. We implement both angle and amplitude encodings of frozen CLIP image embeddings within variational quantum circuits.
In angle encoding, the 512-dimensional embedding is reduced via PCA to 8 dimensions and mapped to rotation angles on 9 qubits in the IQP ansatz. In amplitude encoding, each 512-dimensional CLIP embedding is first normalized to unit norm so that it represents a valid quantum state. The normalized embedding is then directly encoded as the amplitudes of a quantum state
$\ket{\psi} = \sum_{i=1}^{512} x_i \ket{i}.$
To enable the measurement of cosine similarity between image and caption branches, we ensure that image and sentence circuits produce output vectors of matching dimensionality. 

\subsubsection{Collage Encoding}

We also work with a new encoding, which we call {\bf Collage}. This encoding forms a circuit for an image by combining the circuits of each shape in the image. The circuit of each shape comes from its averaged CLIP vector over all single object images of that shape, loaded onto a quantum state, using the two methods of quantum encoding. The two shape circuits are put together either by adding an entangling gate between them, or by juxtaposing them without any connection. The number of qubits in the relational image and sentence circuits are chosen such that they lead to matching output dimensions.

\begin{figure*}[t]
   \centering
   \includegraphics[width=\linewidth]{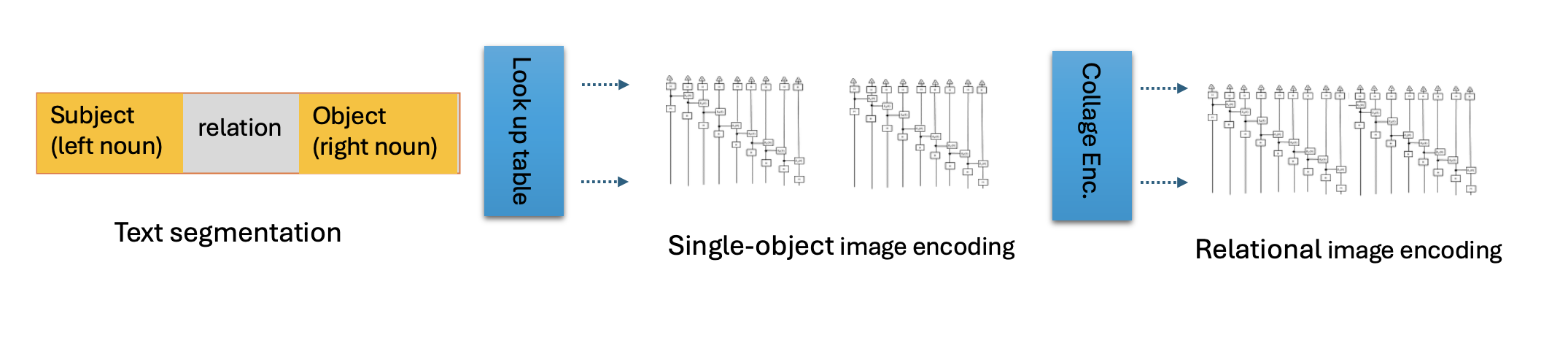}
   \caption{Collage construction of relational image representations from single-object components learned in Stage~1.}
   \label{collage} 
\end{figure*}

\subsection{Caption Encoding and Relation Representation}\label{Caption Encoding}

As stated in section \ref{sec:Disco}, we treat the relational phrases \emph{isLeftOf} and \emph{isRightOf} as a single relational operator with pregroup type $n^{r}s n^{l}$. The relation is encoded by applying the chosen quantum ansatz to the relational box in the DisCoCat diagram, an example of which we gave in Figure \ref{cube-left-cone}. When composed with noun representations on its left and right, the noun wires of the relation contract with their duals, leaving the sentence wire $s$ as the output of the caption diagram. 

In the quantum implementation, each grammatical wire is assigned a fixed number of qubits. The sentence wire $s$ is assigned the same number of qubits as the image representation, ensuring that the caption and image circuits produce output vectors of matching dimensionality. This allows cosine similarity to be computed between the two modalities during training.

\begin{figure*}[t]
   \centering
   \includegraphics[width=0.9\linewidth]{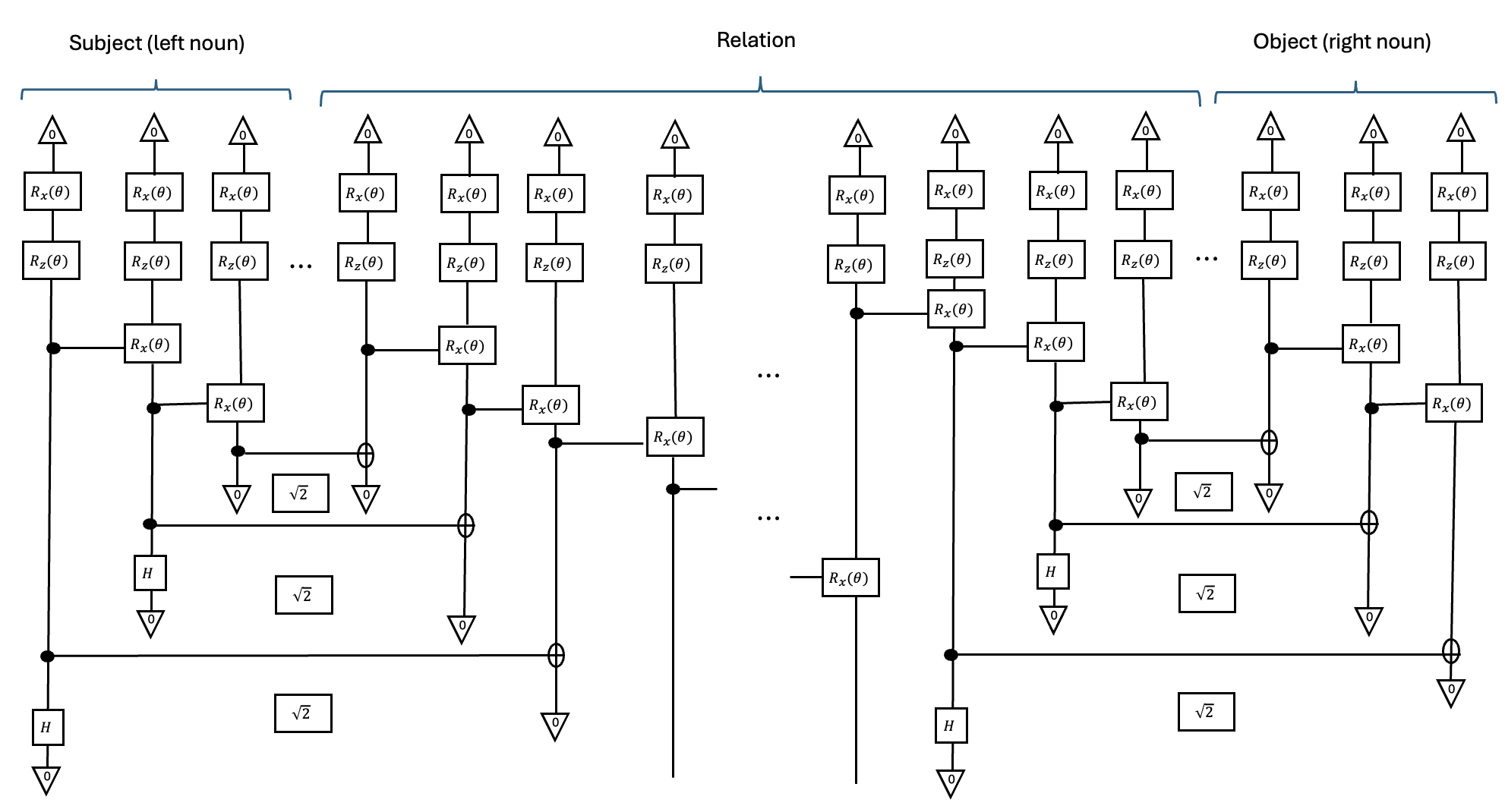}
   \caption{Quantum circuit used for sentence encoding during relational training. Noun parameters are transferred from single-object training and connected to the relational operator. The qubits corresponding to the noun wires are contracted with the relation (via cups) and post-selected, leaving the $s$-wire qubits as the output sentence representation used for similarity computation.}
   \label{text-relation-circuit}
\end{figure*}

Figure~\ref{text-relation-circuit} illustrates the quantum circuit used to encode a relational sentence. The parameters associated with noun components are transferred from Stage~1 and frozen, while the parameters corresponding to the relational part remain trainable.

The outputs of both image and caption circuits are treated as real-valued vectors (obtained from the quantum circuit outputs), and cosine similarity is computed between these vectors during training. We use cosine similarity rather than quantum state overlap (e.g., fidelity) to maintain consistency with classical vision--language models such as CLIP, and to enable a direct comparison between classical and quantum representations within a shared similarity framework.

\section{Experimental Details}
We now describe the training setup for the single-object and relational experiments implemented using the \texttt{lambeq} framework~\cite{kartsaklis2021lambeq}. All variational parameters are randomly initialised and optimised during training.

\subsection{Single-Object Training}
Stage 1 follows the training procedure introduced in Section~\ref{Two-Stage Training Procedure}. Here we describe the optimisation setup used for the single-object experiments. 
We use a contrastive objective based on cosine similarity: the model is trained to assign higher similarity to positive (i.e. matching) image–caption pairs and lower similarity to negative (i.e. mismatched) pairs. In our setting, negatives correspond to images of different shape classes within the same mini-batch. For example, when the caption is \emph{cube}, images of \emph{cone}, \emph{sphere}, and \emph{cylinder} act as negative examples. Negatives are taken from the rest of the mini-batch (“in-batch negatives”), so each step ranks the correct matches against the other candidates without framing the task as binary classification. This setup directly optimises alignment between the two modalities.
Let $B$ be the number of shape classes included per batch and $K$ the number of examples per class. In our experiments we use $B=4$ (cube, cone, cylinder, sphere) and $K=8$.

The contrastive learning setup is summarised below:

\begin{itemize}
\item 
\textbf{Batch size.}
$N=B\times K=4 \times 8=32$

\item
\textbf{Positive and negative examples per caption.}
For each caption in the batch, the model computes cosine similarities between the caption embedding and each image embedding in the same batch. The image that corresponds to the caption forms a positive example, while images belonging to other shape classes act as negative examples:
  \begin{itemize}
    \item
    Positives: $K-1=7$ (the other samples of the same class in the batch)
    \item
    Negatives: $N-K=24$ (all samples from the other classes).
  \end{itemize}

\item
\textbf{Similarity matrix per batch.}
Each forward pass computes cosine similarities between every caption and every image in the batch, forming an $32 \times 32 = 1024 = N \times N$ matrix:
  \begin{itemize}
   \item
     Diagonal (matching pairs): $N=32$
   \item
     Off-diagonal positives: $N \times (K-1)=32 \times 7=224$
    \item
     Off-diagonal negatives: $N \times (N-K)=32 \times 24=768$
  \end{itemize}
\end{itemize}

Supervised contrastive learning works as follows: let $t_i \in \mathbb{R}^d $ be the output of the noun circuit and $v_j \in \mathbb{R}^d$ be the output of the image circuit while $d=512$. We $L2-$normalize both branches, and compute cosine-similarity logits 
$$l_{ij}=\frac{\hat{t_i}^\top \hat{v_j}}{\tau}$$

where $\hat{t_i}$ and $\hat{v_j}$ denote the normalized representations, respectively, and $\tau=0.07$ is the temperature parameter. This value is chosen following common practice in contrastive learning models such as CLIP.

Let $y_i$ denote the class label(shape) of item $i$, and define the set of positives for anchor $i$ as 

$$P(i)= \{j \in \{1,\ldots,N\} \mid y_j = y_i,\, j \neq i \}$$
Thus $|P(i)|= k-1$ in our batches with $k=8$ examples per class. The supervised contrastive loss is :

\begin{equation}
    \label{eq:supcon}
    L_{single}=\frac{-1}{N} \sum_{i=1}^{N} \frac{1}{|P(i)|} \sum_{p \in P(i)} \log \frac{exp(l_{ip})}{\sum_{a=1}^{N} exp(l_{ia})}
\end{equation}
i.e., for each row $i$, we maximize the softmax probability mass assigned to its positive image-caption pairs while pushing down the mass assigned to negative image-caption pairs in that row.
We use this contrastive learning technique for all experiments in Stage 1, namely amplitude encoding and angle encoding of frozen CLIP image embeddings and our classical baseline.

\begin{table*}[t]
\caption{Performance of different ansatz choices for image and text encoding circuits in the single-object training stage. Validation and test splits are out-of-distribution (OOD), and results are averaged over five random seeds.}
    \label{tab:ansatz}
    \centering
    \small
    \renewcommand{\arraystretch}{1.3}
    \setlength{\tabcolsep}{3pt}
    \begin{tabular}{l l c c c c c}
        \toprule
        \textbf{Structure} & \textbf{Train} & \textbf{Valid} & \textbf{Test} & \textbf{parameters} & \textbf{Text-qubits} & \textbf{Image-qubits}\\
        \midrule
        \textbf{Image\_sim4 \& Text\_sim4} & 24.76\% & 25\% & 33.33\% & 96 & 3 & 3\\
        \midrule
        \textbf{Image\_sim4 \& Text\_IQP} & 27.18\% & 25\% & 33.33\% & 80 & 3 & 3\\
        \midrule
        \textbf{Image\_IQP \& Text\_IQP} & 80.54\% & 45\% & 53.34\% & 320 & 9 & 9 \\
        \midrule
        \textbf{Image\_IQP \& Text\_sim4} & 78.66\% & 75.00\% & 66.70\% & 312 & 9 & 9\\
        \bottomrule
    \end{tabular}
\end{table*}

\subsubsection{Quantum Circuit Implementation Details}\label{Quantum-Circuit-Implementation-Details}
\begin{figure}[t]
   \centering
   \includegraphics[width=\linewidth]
    {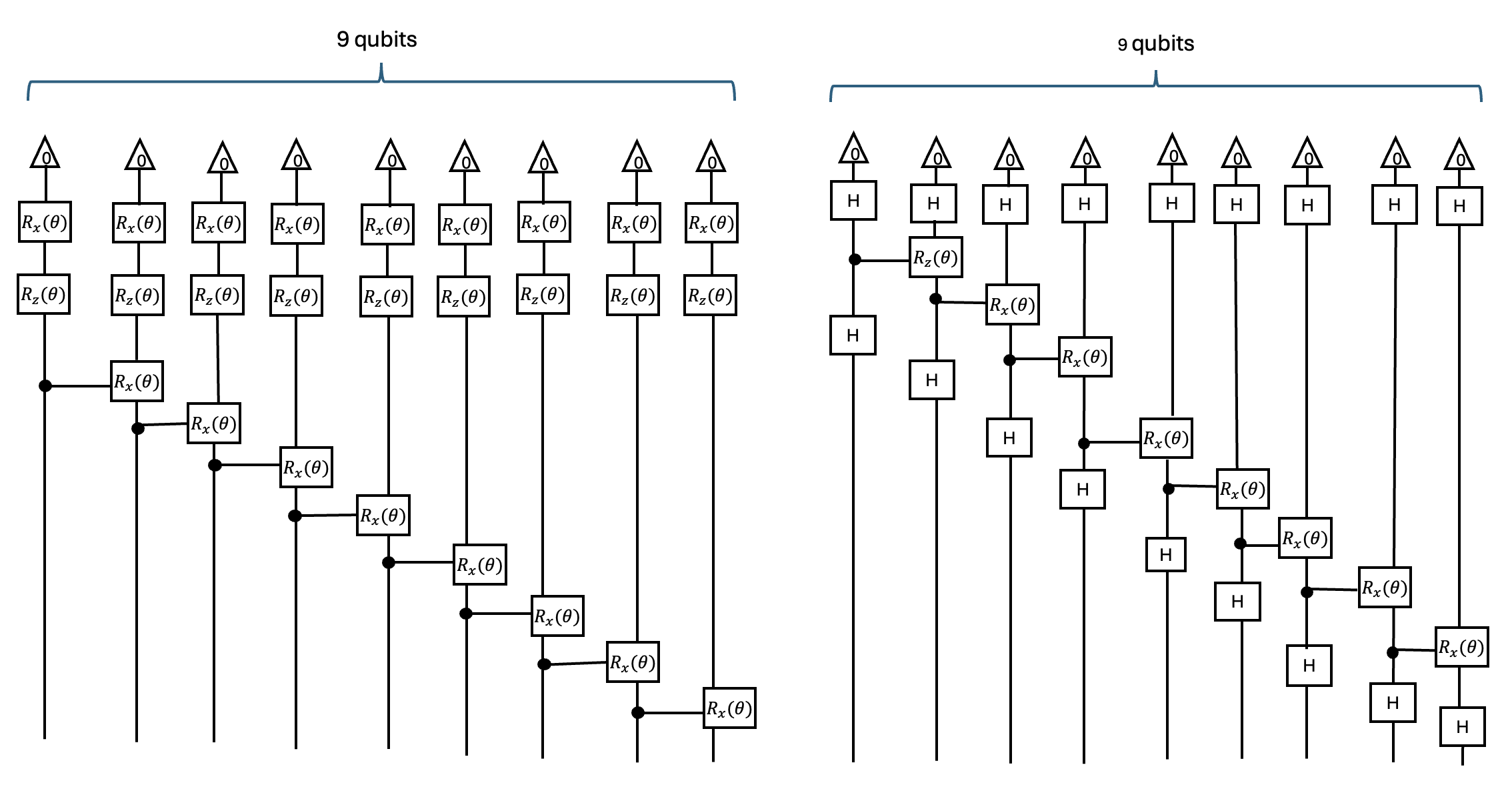}
   \caption{The left circuit depicts the quantum circuit used for sentence encoding, while the right circuit depicts the quantum circuit used for image encoding during single-object training.}
   \label{single-object-circuits}
\end{figure}

For angle encoding in single-object stage, we explore different ansatz choices for both the sentence and image encoders. Table~\ref{tab:ansatz} reports the number of qubits, trainable parameters, and resulting accuracies obtained using the Sim4 and IQP ansatzes.
For encoding the 8-dimensional PCA-reduced image embeddings, the IQP and Sim4 ansatzes require 9 and 3 qubits, respectively. To ensure compatibility between image and noun representations when computing cosine similarity, we use the same number of qubits for the noun circuit.
To balance expressivity and circuit complexity, we compare these ansatz choices and their entangling capabilities across modalities. Table \ref{tab:ansatz} shows that the  the IQP ansatz for image encoding and the Sim4 ansatz for noun circuits is most effective. 
The number of qubits required by each ansatz is determined by the encoding scheme rather than by the dimensionality of the input alone. In our setting, both IQP and Sim4 encode the same low-dimensional PCA features, but differ in how these features are mapped into quantum circuits. IQP encoding distributes input features across a larger number of qubits through its structured circuit design, whereas Sim4 encodes the same features using a more compact qubit register with a different arrangement of parameterised and controlled operations. Therefore, differences in performance cannot be attributed solely to qubit count, but instead reflect differences in encoding strategy and circuit structure.
These ansatzes are used in our main experiments reported in Tables~\ref{tab:sinle-object} and~\ref{tab:relational}.
Figure \ref{single-object-circuits} shows quantum circuits of image and noun in single-object training. 

\subsection{Relational Training}

Stage 2 follows the relational alignment setting introduced in Section~\ref{Methods}. Here we describe the optimisation objective used for relational training.

The model is trained to assign higher similarity between the image and the correct caption that between the image and the distractor caption.
The model computes cosine similarity between the outputs of the caption and image representations and is trained using a margin-based ranking loss:
\begin{equation}
L_{rel} = \tfrac{1}{2} \big( y(1 - s)^2 + (1 - y)\max(0, s - m)^2 \big),
\end{equation}
where $y \in \{0,1\}$ denotes the ground-truth label, $s$ is the cosine similarity, and $m$ is a margin (set to $0.5$ for quantum experiments). This objective differs from the supervised contrastive loss used in Stage 1, because the relational task is formulated as a \textit{pairwise} comparison between a correct caption and a relation-swapped distractor. 
Since we are contrasting only two items, the supervised contrastive loss used for the single objective stage (Equation \eqref{eq:supcon}) is less appropriate for this task.
We evaluate relational learning across multiple image encoding strategies:

\begin{enumerate}
    \item 
    \textbf{Multi-hot encoding (MHE) for images:} relational images are represented using multi-hot encodings, as explained in Section~\ref{MHE}. Caption encoders reuse frozen noun parameters obtained from single-object training, while relation parameters remain trainable. Details of the caption encoding are provided in Section~\ref{Caption Encoding}.
    \item \textbf{Angle encoding of CLIP image embeddings}: relational images are encoded using angle encoding of frozen CLIP image embeddings. Caption encoders follow the staged freezing scheme described above, where noun parameters are transferred from Stage~1 and frozen, and only relational parameters are optimised.
    \item 
    \textbf{Amplitude encoding of CLIP image embeddings:} relational images are encoded using amplitude encoding of frozen CLIP image embeddings. As in the previous cases, caption noun parameters are transferred from Stage~1 and frozen, while only relational parameters are trained.
    \item
    \textbf{Classical baseline}: relational captions are trained using fixed CLIP image representations, with cosine similarity computed directly between classical CLIP embeddings. This serves as a direct comparison point to evaluate the benefit of quantum representations.
    \item
    \textbf{Collage encoding with angle encoding.}
    Relational images are constructed by composing the quantum circuits of the two constituent shapes transferred from single-object training. Each shape is represented by a 9-qubit circuit; therefore, the relational image circuit has 18 qubits. The relational image circuit is formed either by introducing an entangling gate between the two shape circuits in the IQP ansatz or by juxtaposing them without interaction, resulting in an 18-qubit image circuit. To enable cosine similarity computation, relational captions are encoded using sentence circuits with a matching number of qubits, i.e., 18.
    \item 
    \textbf{Collage encoding with amplitude encoding.}
    Relational images are constructed by composing amplitude-encoded quantum states transferred from single-object training. We consider two variants: a unified-state representation using 10 qubits to encode 1,024 amplitudes, and a multiple-state representation using two 9-qubit states (one per shape), yielding 18 qubits in total.
\end{enumerate}
Our models are named as follows. For single-object images (stage one), we have \textbf{OHE}:one-hot quantum encoded images; \textbf{Q-CLIP}: quantum encoded CLIP images; \textbf{Classical-CLIP}: classically encoded CLIP images. For relational images (stage 2), we have \textbf{MHE}: multi-hot quantum encoded images; \textbf{Collage Encoding}: Collage quantum encoded images; \textbf{Q-Multi-CLIP}: multistage training on quantum encodings of CLIP images. For comparison, we also implement a single-stage training baseline \textbf{Q-Single-CLIP}, in which we attempt to learn caption embeddings directly from the relational images, and \textbf{Classic-Rel-CLIP} where we use classical encodings of CLIP images and text captions. All models are trained using the Adam optimizer for up to 50 epochs with early stopping based on validation performance. 
Learning rates are selected by evaluating a small set of candidate values on the validation set and choosing the value that yields the best validation performance. 
The results are averaged over five runs with different random seeds. Batch size is set to 16 for angle encoding and multi-hot encoding, and to 48 for amplitude encoding. Relational models are implemented using the lambeq framework, specifically its \texttt{PytorchQuantumModel} interface.

\section{Results and Analysis}

The experimental results are presented progressively. Table~\ref{tab:ansatz} first compares candidate ansatz configurations for the single-object stage and motivates the architecture used in the subsequent experiments. Table ~\ref{tab:sinle-object} then evaluates the selected models on the single-object task, establishing the quality of the learned object representations. Finally, Table~\ref{tab:relational} reports the relational learning results obtained by transferring these learned object representations to the second stage, allowing us to assess compositional generalization.

\begin{table*}[t]
\caption{Results of single object training. Columns Train, OOD-Valid, and OOD-Test reported average accuracy over 5 random seeds. Best Test reports the best accuracy achieved across the 5 seeds. Parameters reports number of parameters of each model.}
    \label{tab:sinle-object}
    \centering
    \renewcommand{\arraystretch}{1.2}
    \setlength{\tabcolsep}{3pt}
    \begin{tabular}{l l c c c c c}
        \toprule
        \textbf{Models} & \textbf{Method} & \textbf{Train} & \textbf{OOD-Valid} & \textbf{OOD-Test} & \textbf{Best Test}& \textbf{Parameters}\\
        \midrule
        \textbf{OHE} &  & 92.20\% & 95.00\% & 93.34\% & 100.00\% & 168\\
        \midrule
        \textbf{Q-CLIP} & Angle Enc. & 78.66\% & 75.00\% & 66.70\% & 70.31\% & 312\\
                              & Amplitude Enc. &87.37\% &79.71\% & 80.97\% & 100.00\% & 312\\
         \midrule
        \textbf{Classical Baseline}\\
         \midrule
         \textbf{Classical-CLIP} & CTP & 95.63\% & 90.45\% & 91.00\% & 95.18\% & 63,428,097\\
        \bottomrule
    \end{tabular}
\end{table*}

Results are presented in Table~\ref{tab:sinle-object} and Table~\ref{tab:relational}, reported as average test accuracy over five random seeds, along with the corresponding best test accuracy across seeds. Hyperparameters are selected based on validation performance, and validation data is used for model selection and early stopping.
In the single-object experiments, we see that the OHE proof-of-concept implementation achieves 93.34\% accuracy on the OOD test set. This shows that with with fully separated image embeddings, we are able to effectively learn corresponding embeddings for their textual labels. 

Among the models trained with quantum encodings of CLIP image vectors (\textbf{Q-CLIP}), Amplitude Encoding performs strongest, 
achieving accuracy of 80.97\%. Although achieving lower accuracy than the classical baseline (91\%), our model uses only a fraction of the trainable parameters: 312 compared to approximately 63M in the classical model. We note that this comparison considers only the trainable components of each model. In both the classical and quantum settings, CLIP image embeddings are kept fixed, and only the text-side parameters are optimised. Therefore, the reported parameter counts reflect the effective number of trainable parameters rather than the total size of the pretrained CLIP model. 

\begin{table*}[t]
\caption{Results of relational learning.Columns Train, OOD-Valid, and OOD-Test reported average accuracy over 5 random seeds. Best Test reports the best accuracy achieved across the 5 seeds. Parameters reports number of parameters of each model.}
    \label{tab:relational}
    \centering
    \renewcommand{\arraystretch}{1.3}
    \setlength{\tabcolsep}{3pt}
    \begin{tabular}{l l c c c c c}
        \toprule
        \textbf{Models} & \textit{Method} & \textbf{Train} & \textbf{OOD-Valid} & \textbf{OOD-Test} & \textbf{Best Test}& \textbf{Parameters} \\
        \midrule
        {\bf MHE} & Binary & 99.00\% & 73.00\% & 85.88\% & 91.66 & 279\\
        \midrule
        \bf Collage Encoding & \\
         & Angle Enc. with Entangl. & 82.20\% & 61.05\% & 72.32\% & 75.00 & 426\\
          & Angle Enc. No  Entangl.   & 82.11\% & 64.00\% & 71.66\% & 75.00 & 426\\
          & Ampl. Enc. with Entangl.& 62.28\% & 40.00\% & 48.58\% & 50.00 & 426\\
          & Ampl. Enc. No Entangl. & 60.14\% & 40.00\% & 43.33\% & 50.00 & 426\\
        \midrule
        \textbf{Q-Multi-CLIP} & Angle Enc. & 76.60\% & 61.25\% & 55.33\% & 60.16 & 480\\
                               & Amplitude Enc. & 59.02\% & 45.00\% & 50.00\% & 50.00 & 480 \\
           \midrule
           \textbf{Baselines} 
          \\
          \midrule
          {\bf Q-Single-CLIP} &  Angle Enc. & 55.26\% & 44.50\% & 53.09\% & 55.00 & 792 \\
         & Amplitude Enc.   & 50.64\% & 62.00\% & 42.93\% & 48.33 & 792 \\
           \midrule
          \textbf{Classic-Rel-CLIP} & CTP & 75.94\% & 11.14\% & 50.00\% & 53.20 & 151,540,225\\
        \bottomrule
    \end{tabular}
\end{table*}




In the quantum models, Q-Multi-CLIP reuses the noun parameters learned during Stage~1, whereas Q-Single-CLIP is trained directly on the relational task without parameter transfer, providing a single-stage baseline for evaluating the benefit of the proposed multi-stage training strategy.

We see that the proof-of-concept \textbf{MHE} encoding performs strongly, with almost perfect training accuracy and an average OOD test accuracy of 85.88\%. We further see that our \textbf{Collage} encoding performs strongly when paired with angle encoding of the image embeddings, obtaining 72\% accuracy, substantially higher than random guessing and the single-stage training baseline (\textbf{Q-Single-CLIP}), while using substantially fewer parameters than the classical baseline. Finally, directly encoding the frozen CLIP image embeddings (\textbf{Q-Multi-CLIP}) via angle and amplitude encoding performs less strongly (55.33\% and 50\%, respectively).

For comparison, the classical baseline (\textbf{Classic-Rel-CLIP}) exhibits limited out-of-distribution generalization. Although it achieves 75.94\% training accuracy, its OOD test accuracy drops to 50\%, indicating that it fails to generalize to unseen relational compositions.

We note that OOD validation accuracy can be lower than OOD test accuracy due to differences in the difficulty of the held-out compositional splits, which arise from the specific partitioning of compositions between validation and test sets.

\subsection{Analysis}
\begin{figure*}[!t]
    \centering
    \begin{subfigure}{0.35\textwidth}
        \centering
        \includegraphics[width=\linewidth]{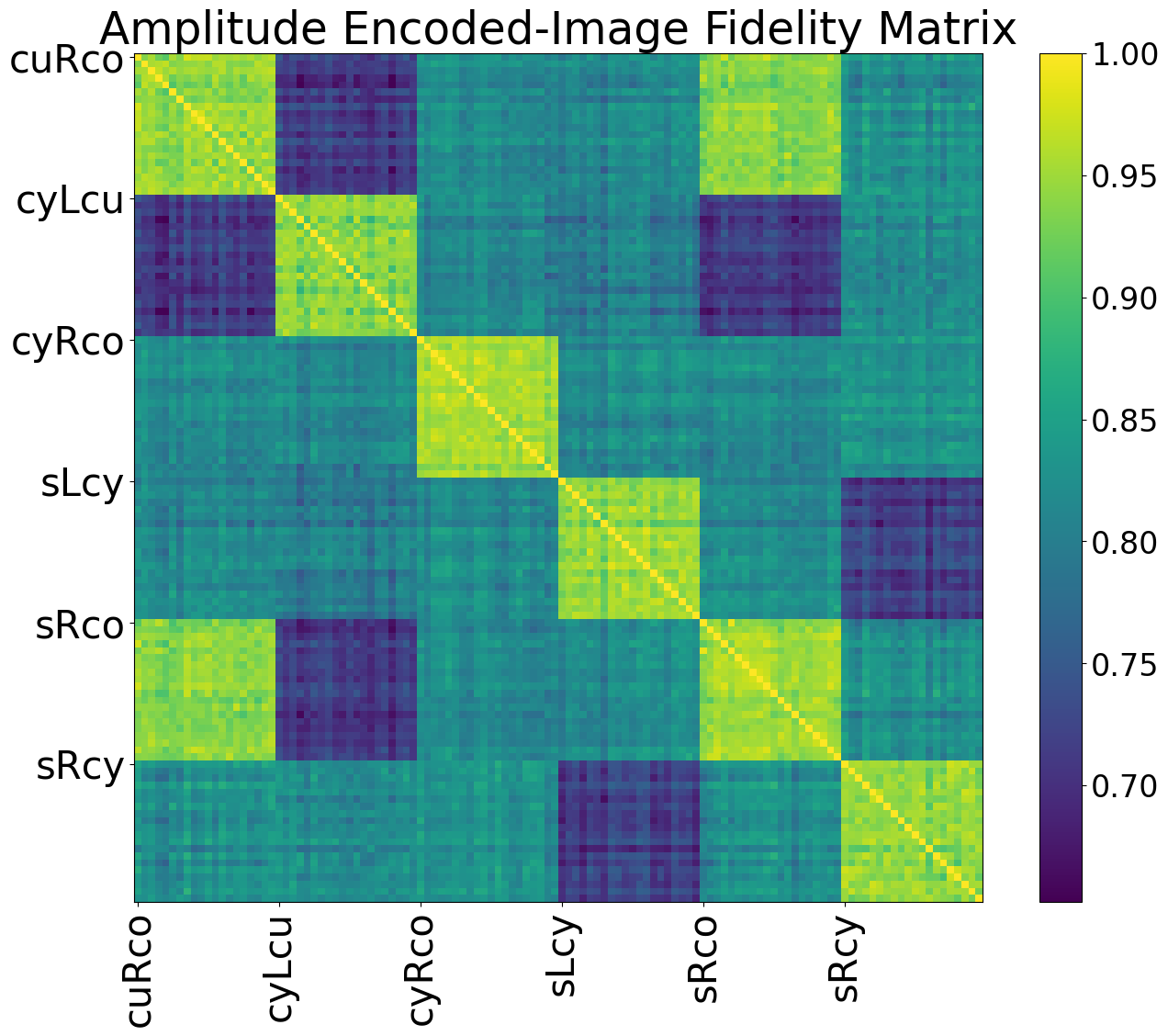}
        \caption{Amplitude encoding}
        \label{fig:1a}
    \end{subfigure}
    \hspace{0.5cm}
    \begin{subfigure}{0.35\textwidth}
        \centering
        \includegraphics[width=\linewidth]{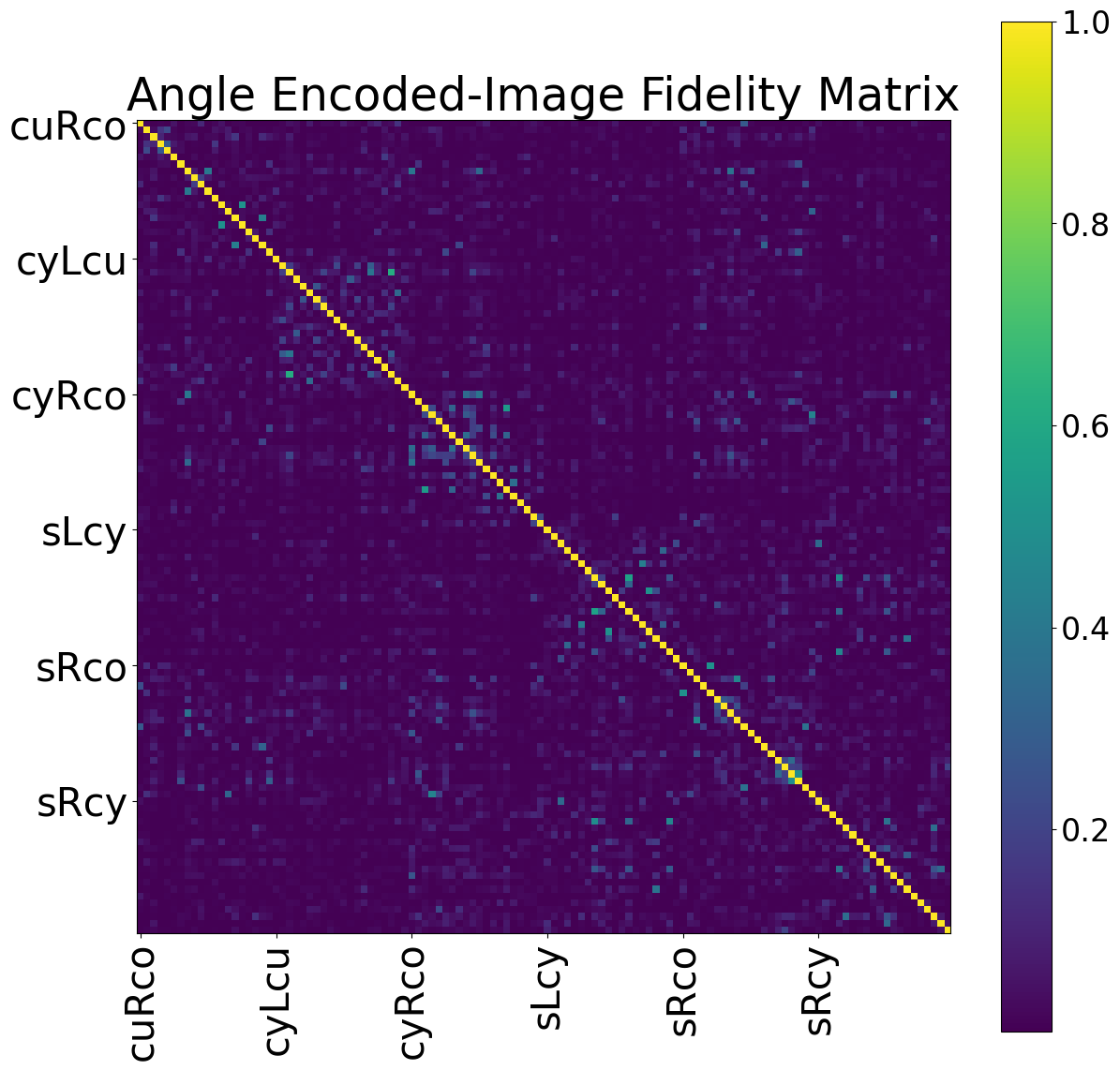}
        \caption{Angle encoding}
        \label{fig:1b}
    \end{subfigure}
    \caption{Amplitude vs. angle encoding state fidelity on the test images. The desired pattern is block diagonal, indicating high within-class similarity and low between-class similarity. Labels use the abbreviations: cu (cube), cy (cylinder), co (cone), s (sphere), L (left), and R (right).}
    \label{angle-amplitude}
\end{figure*}

To understand the less strong performance of \textbf{Q-Multi-CLIP}, we plot a representational similarity analysis (Figure~\ref{angle-amplitude}). This gives a heatmap of the similarities, as measured by state fidelity, of the quantum encodings of CLIP image embeddings. The desired pattern in these plots is a block diagonal: between  all images of a particular class, e.g. \textit{cube left cone}, similarity should be high, and between images of two different classes, similarity should be low.

In Figure~\ref{angle-amplitude} we do not see this pattern for either amplitude or angle encoding. For amplitude encoding (Figure \ref{fig:1a}), we see that while the block diagonal pattern is somewhat present, there are also off-diagonal blocks, as well as overall high similarity across all quantum encodings (note the heatmap scale). For angle encoding (Figure \ref{fig:1b}), we do not see a block diagonal, indicating that similarity of quantum encodings is low both within and outside a given class. The diagonal line indicates that only self-similarity of a given image encoding is high. We note that the similarity matrices shown in Figure~\ref{angle-amplitude} correspond to the quantum state representations after encoding. While one could also examine similarity structures in the underlying classical CLIP embeddings (with and without PCA), these are not expected to exhibit strong compositional structure, as CLIP is not explicitly trained for relational reasoning. The improvements observed with angle encoding therefore arise from the encoding transformation itself, rather than from properties of the original embedding space.

An important question is whether compositional sub-structures are already present in the underlying CLIP embeddings. While CLIP captures strong semantic similarity at the object level, prior work has shown that it does not reliably encode compositional relationships between objects and relations. This is consistent with our observations, where classical CLIP-based baselines fail to generalise to unseen relational compositions. The role of the quantum embedding can therefore be interpreted as enhancing or restructuring the representation space to better reflect compositional structure, rather than simply extracting it from the original embeddings.

We suspect that these patterns arise because angle encoding introduces nonlinear feature transformations (via sinusoidal rotations) that increase the separability of spatial relations, whereas amplitude encoding preserves the original embedding geometry of CLIP, which does not explicitly encode relational structure. As a result, amplitude-encoded representations of \emph{left} and \emph{right} remain highly similar. 
The absence of the required block-diagonal patterns indicates that the CLIP image embeddings are themselves inadequate to distinguish the required spatial relations. 
\section{Conclusion and Future Work}
We presented a quantum multimodal framework for compositional concept generalization based on DisCoCat and variational quantum circuits. By introducing a multi-stage training procedure that separates object grounding from relational learning, we demonstrate improved generalization and parameter efficiency compared to classical baselines. The strongest relational performance arises from structured encodings that preserve compositional structure, particularly the collage construction combined with angle encoding. These results highlight the importance of structured representation and staged learning for compositional generalization in grounded vision–language models. All code used in this work, including implementations of single-object and relational training, image--caption alignment, and multi-stage parameter transfer, is publicly available at
\href{https://github.com/Mina-Abbaszade/Quantum-Multimodality}{https://github.com/Mina-Abbaszade/Quantum-Multimodality}.

This work leads to two  lines of  future research 
\begin{itemize}
\item Our experiments are conducted in simulation to isolate the representational and compositional properties of the proposed framework in a controlled setting. This allows us to study the effect of compositional structure independently of hardware noise, which is standard practice in quantum machine learning when evaluating model design. The circuits used in this work are based on standard variational ansatz (e.g., IQP and Sim families) and operate on average qubit counts (e.g., 9 qubits for single-object representations and up to 18 qubits for relational encodings). Making these compatible with near-term quantum devices by reducing qubit count and depth,  implementing the circuits on available hardware, and evaluating  performance under realistic noise  is an important direction for future work.

\item Our experiments focus on a controlled benchmark with a small number of objects and spatial relations. Previous work has shown that even such minimal compositional tasks remain challenging for modern vision--language models, particularly in out-of-distribution generalization \cite{hsieh2023sugarcrepe, thrush2022winoground, hupkes2020compositionality}.  An important direction is to better understand how such structured approaches behave as compositional complexity increases. In particular, extending these methods to richer relational vocabularies and more complex scenes is not straightforward, and requires careful consideration of both representation and training dynamics. 

\end{itemize}


\bibliographystyle{IEEEtran}
\bibliography{refs}

\end{document}